\documentclass{article} 
\usepackage{iclr2027_conference,times}

\usepackage[utf8]{inputenc} 
\usepackage[T1]{fontenc}    
\usepackage{hyperref}       
\usepackage{url}            
\usepackage{algorithm}
\usepackage{algpseudocode}
\usepackage{booktabs}       
\usepackage{amsfonts}       
\usepackage{amsmath}        
\usepackage{amssymb}        
\usepackage{mathtools}      

\renewcommand{\cite}[1]{{\color{red}USE CITEP OR CITET}}

\newcommand{\mb}{\mathbb}
\newcommand{\R}{\mb R}
\newcommand{\E}{\mathbb{E}}
\newcommand{\defeq}{\ensuremath{\stackrel{\rm def}{=}}}

\newcommand{\obj}{\mathcal{O}}
\newcommand{\family}{\mathcal{F}}
\newcommand{\sampler}{\mathcal{S}}
\newcommand{\sampleset}{\mathcal{D}}
\newcommand{\sample}{s}
\newcommand{\weight}{w}
\newcommand{\decoder}{g_\psi}
\newcommand{\lossfn}{\mathcal{L}}
\newcommand{\embedding}{\theta}
\newcommand{\embedder}{\mathcal{E}_\phi}

\newcommand{\reals}{\mathbb{R}}
\newcommand{\prior}{p_0}
\newcommand{\state}{x}

\title{Universal Sample-Based Embeddings\\of Higher-Order Objects}

\author{%
Radovan Haluška\textsuperscript{1},
Christopher Solinas\textsuperscript{2},\\
\textbf{Nathan R. Sturtevant\textsuperscript{2,3},
Martin Schmid\textsuperscript{1,4},
Michael Bowling\textsuperscript{2,3}}
\\
\textsuperscript{1}Charles University, Prague,
\textsuperscript{2}University of Alberta,\\
\textsuperscript{3}Alberta Machine Intelligence Institute (Amii),
\textsuperscript{4}EquiLibre Technologies, Inc.
\\
\texttt{\{haluskr,schmidm\}@kam.mff.cuni.cz},\\
\texttt{\{solinas,nathanst,mbowling\}@ualberta.ca}
}

\iclrfinalcopy 
\begin{document}

\maketitle

\begin{abstract}
Neural networks operate naturally on finite-dimensional vectors, yet many problems require reasoning about higher-order objects such as distributions, functions, and graphs. Such objects capture relationships within a ground set and rarely admit an explicit compact representation. Moreover, we typically want a compact representation not of a single higher-order object but of an entire family of such objects, including members unseen during training. While a few approaches exist for particular object types, none is designed to work across object types. We propose a unified framework that represents any higher-order object as a fixed-size embedding vector learned over a family of such objects. An encoder maps an object, seen only through a finite set of samples, to an embedding, and a domain-specific decoder supplies the training signal by modeling the object conditioned on that embedding. Since the encoder works only from samples, one architecture serves different object types. Because the decoder is a model already suited to single instances of the object type, the framework inherits that model's native operations and inductive biases. We instantiate the framework on distributions, functions, and graphs. In each case we show that a single trained model represents an entire family and generalizes to members unseen during training. Furthermore, we demonstrate the downstream utility of these embeddings: interpolating distributions in embedding space, solving differential equations in embedding space, and predicting unseen edges from a graph's embedding.
\end{abstract}

\section{Introduction}
\label{sec:introduction}

Neural networks have seen a widespread adoption in practically every field of science and engineering where inputs and outputs can be represented as finite-dimensional vectors. In such cases, existing approaches excel at approximating complex relationships between inputs and outputs coming from complex distributions \citep{lecun2015deep, goodfellow2016deep, wang2023scientific}. However, there is no natural way of \emph{capturing} distributions, functions, or generally any higher-order objects, as inputs or \emph{producing} them as outputs. For instance, it is not obvious how to train a neural network to predict the probability density function of an input distribution as there is no general compact representation to provide to the network.

Once we allow higher-order objects as inputs and outputs, it is natural to restrict ourselves to the relevant \emph{families} of such objects. Examples include families of distributions that share a common parameterization, such as the family of Gaussian distributions parameterized by their mean and covariance, or solutions to differential equations under varying parameters and initial conditions. When a family admits a closed-form parameterization, one can simply feed the parameters into the neural network, but in many cases such a parameterization is not known a priori. Manipulating these higher-order objects, and families thereof, efficiently, with generalization to unseen members, has no general and established solution and is typically handled in ad-hoc and domain-specific manners \citep[e.g.,][]{edwards2016statistician, kovachki2023neuraloperator, narayanan2017graph2vec}.

These approaches each target a single object type and carry inductive biases that do not transfer to the others. For distributions, kernel mean embeddings \citep{muandet2017kme} depend on a hand-crafted kernel, and the Neural Statistician \citep{edwards2016statistician} ties its embeddings to a variational autoencoder. For functions, Neural Processes \citep{garnelo2018neuralprocesses} learn distributions over functions, and neural operators such as DeepONet \citep{lu2021deeponet} and FNO \citep{li2020fno} learn maps between function spaces, but impose rigid input assumptions and expose no representation of an individual function. For graphs, most methods embed nodes rather than whole graphs \citep{hamilton2017graphsage} or learn to generate graphs \citep{simonovsky2018graphvae, you2018graphrnn, vignac2022digress} without exposing an embedding of any single one. More recently, \emph{distribution embeddings} have been proposed independently in two distinct domains: as part of a filtering algorithm for state-space models \citep{solinas2026nbf} and as a way to reason across scales in computational biology \citep{fishman2026gde}.

We propose a unified framework for representing and learning from families of higher-order objects, building on the distribution embeddings introduced in  the initial version of \citet{solinas2026nbf}. It couples an \emph{encoder}, which maps an object given only as a finite set of samples to a fixed-size vector, with a domain-specific \emph{decoder} that provides the training signal. Because the encoder sees only samples, a single architecture handles very different object types; because the decoder is the model one would normally fit to a single object, the framework lifts its native operations, such as sampling, density evaluation, and spectral reconstruction, from one object to a whole family, while producing a compact representation of each member that is reusable in downstream tasks.

We instantiate the framework on three object types, changing primarily the decoder. For \emph{distributions}, a generative decoder reconstructs densities from few samples, and the embedding space supports arithmetic and interpolation between distributions. For \emph{functions}, a pointwise decoder reconstructs functions faithfully and scales to high-dimensional state; because whole functions become vectors, an ordinary network acts as a neural operator, which we use to differentiate functions and solve ODEs in embedding space. For \emph{graphs}, a spectral decoder recovers a Laplacian representation that reliably separates edges from non-edges, from sampled edges alone. In every case, a single trained model generalizes to unseen family members and to sample sets of arbitrary size.

\section{Embedding \& Learning From Families of Higher-Order Objects}
\label{sec:object_embeddings}

For any given higher-order object, there is usually a well-established way of approximating it --- for instance, generative models for distributions \citep{kingma2013vae, papamakarios2021nfs, lipman2022fm}. Such approaches, however, treat every object in isolation: whenever the target changes, the model must be retrained from scratch, and it never yields an explicit compact representation of the object. Our framework instead replaces per-object fitting with a single model that represents a whole family of objects and generalizes to members unseen during training.

Our goal is two-fold: a) to learn representations of higher-order objects, and b) to enable learning across families of such objects. We view a \emph{higher-order object} $\obj$ as anything that has no canonical finite-vector representation, such as a distribution, a function, or a graph. A \emph{family} $\family$ is a collection of related objects; for example, the solutions of a given differential equation across its initial conditions and parameters form a family.

For a family of higher-order objects to be learnable within our framework, we require
the following three components:

\begin{enumerate}
    \item \textbf{Object sampler.} A procedure $\sampler$ that, given an object
    $\obj \in \family$, produces a set of (weighted) samples $\sampleset$. This set, if sufficiently large, represents
    $\obj$ and is our only access to it; we never require a closed-form description
    of the object.
    \item \textbf{Parameterizable fitting model.} A model $\decoder$ that can learn
    to represent an object $\obj$ from a set of samples $\sampleset$.
    \item \textbf{Training loss.} A loss function $\lossfn$ that provides the only
    training signal for the framework. It is usually tied to the fitting model $\decoder$.
\end{enumerate}

\begin{figure}[t]
    \centering
    \includegraphics[width=\textwidth]{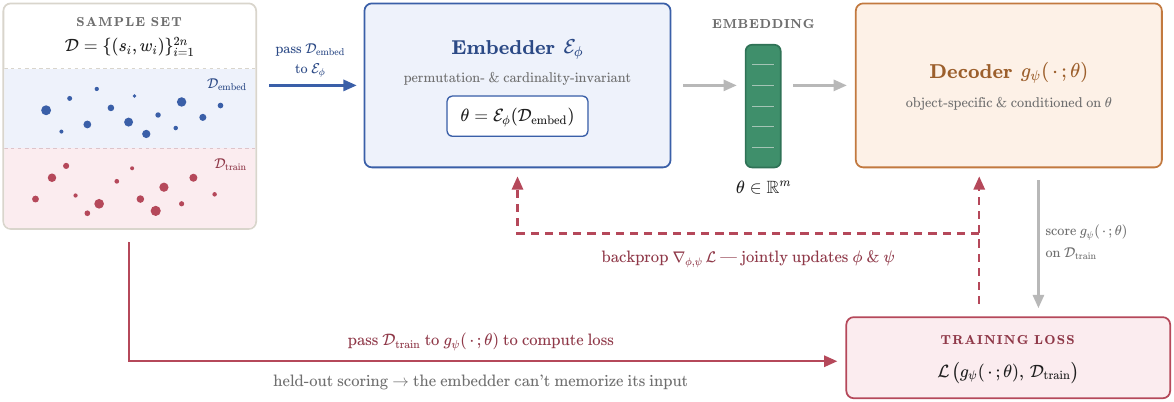}
    \caption{%
        \textbf{General training procedure.} Each sample set $\sampleset$ is split
        into two halves, $\sampleset_\text{embed}$ and $\sampleset_\text{train}$. The
        embedder $\embedder$ maps $\sampleset_\text{embed}$ to an embedding vector
        $\embedding$ that conditions the object-specific decoder $\decoder$. The held-out
        half $\sampleset_{\text{train}}$ is then used to score $\decoder$ and compute
        the training loss $\lossfn$. Finally, $\lossfn$ is differentiated with respect
        to the trainable parameters of both networks and backpropagated.%
    }
    \label{fig:training_procedure}
\end{figure}

To make the three components concrete, consider a family of two-dimensional, donut-shaped distributions. Each object $\obj$ is a single donut, fully described by three parameters: a mean, a radius, and a width, although this parameterization is not known a priori. The family $\family$ is itself a distribution over such donuts. A sample set $\sampleset$ for a given donut is a collection of i.i.d. points $\{x_i\}_{i=1}^n$, $x_i \in \R^2$, drawn from it; Figure~\ref{fig:distribution_embeddings} shows several donuts, each represented by its own $\sampleset$. The fitting model $\decoder$ can be any generative model, such as a Normalizing Flow, and the training loss is the negative log-likelihood of the samples under that model.

Given $\sampler$, $\decoder$, and $\lossfn$, we train a single model that a) learns
to represent the objects in a family as compact embedding vectors $\embedding \in
\R^m$, from only a partial view of each object through samples, and b) learns to
fit an object given its embedding and an object-specific input. To this end, our
model comprises two jointly-trained neural networks: an \emph{embedder}
$\embedder$, sometimes called an \emph{encoder}, and the supplied fitting model $\decoder$, which we will refer to as
the \emph{decoder}.

The role of the embedder is to compress an object's samples into a fixed-size
vector that retains the information needed to fit the object. Using sets of
samples as the only input has two benefits. First, it lets a single architecture
serve very different object types, as the following sections show. Second,
it lets us handle objects seen only through samples, at arbitrary sample-set
sizes, including objects never encountered during training. The fixed-size
embedding is convenient in turn: it can serve as a \emph{conditioning signal}, or
as an input to and output of other networks. Our framework conditions the decoder
$\decoder$ on the embedding; we explore further uses, such as operator learning,
in later sections.

On the other hand, because the decoder is what we would normally use to fit a
single object, it not only drives training by providing the loss signal, but also
lets the whole model inherit the decoder's native capabilities, such as sampling
and density estimation in Flow Matching models. In this way the framework extends
object-specific operations, normally available on a per-object basis, to entire
families, while also producing representations suitable for downstream tasks.

\paragraph{Training data.} Our training data is a finite collection of sample sets
$\{\sampleset_1, \sampleset_2, \dots, \sampleset_N\}$, where each $\sampleset_k = \{(\sample_i, \weight_i)\}_{i=1}^{2n}$ is
a sample-based representation of an object $\obj_k$, collected with the object
sampler $\sampler$. We do not require this mapping to be unique: a single $\obj_k$
may be described by several overlapping sample sets, possibly of varying size, and we
make no assumptions about their structure, such as a regular grid of points.
Working only with samples, rather than object-specific encodings, is precisely
what lets our framework support families of very different objects. The training
data can therefore be viewed as a \emph{dataset of datasets}: each $\sampleset_k$
describes a single object and is itself an ordinary dataset, though typically a
small one. Equivalently, a family is a distribution over datasets, and our model
learns to fit this distribution and to generalize across it.

\paragraph{Embedding network.} The embedder takes a sample set $\sampleset$
representing a member of the family and produces a fixed-size embedding vector
$\embedding \in \R^m$. The network must be \emph{cardinality}- and
\emph{permutation}-invariant: its output must not depend on the size of the set or
on the order of its elements. DeepSets \citep{zaheer2017deepsets} and Set
Transformers \citep{lee2019st} are examples of such architectures. The embedder
encodes each sample separately, possibly conditioned on the rest of the set, and pools the resulting codes by a (weighted) mean into a single vector. In
Section~\ref{sec:distribution_embeddings}, we discuss a variation on this
architecture that better captures higher moments.

\paragraph{Training procedure.} In each training step we sample a batch of $B$
sample sets $\{\sampleset_1, \sampleset_2, \dots, \sampleset_B\}$, each of size
$2n$. Every sample set $\sampleset_k$ is split into two halves of approximately
equal total weight. The first half $\sampleset_{k, \text{embed}}$ is passed to
the embedder to produce an embedding $\embedding_k = \embedder(\sampleset_{k,
\text{embed}})$, and the second half $\sampleset_{k, \text{train}}$ is scored by
the decoder $\decoder$ conditioned on $\embedding_k$. This split discourages the
embedder from memorizing its input and forces $\embedding_k$ to capture
transferable structure of $\obj_k$. The loss $\lossfn$ itself is
object-dependent, and we discuss several decoders and their loss functions in the
following sections. In every case, however, the training signal is backpropagated
through the decoder and the embedding $\embedding_k$ back to the embedder, so
that the two networks are trained jointly and the embedder learns to produce
embeddings the decoder can use. This procedure is illustrated in
Figure~\ref{fig:training_procedure}.

In the following three sections, we instantiate and evaluate our approach on concrete
families of distributions (Sec.~\ref{sec:distribution_embeddings}), functions
(Sec.~\ref{sec:function_embeddings}), and graphs (Sec.~\ref{sec:graph_embeddings}). The framework is
not limited to these: any higher-order object and family thereof fits within it, given the three
components introduced at the start of this section.

\section{Distribution Embeddings}
\label{sec:distribution_embeddings}

We first instantiate the framework on families of distributions. Every distribution
in a family shares a common sample space, and a single sample is one i.i.d. draw
from a member distribution.

Following Section~\ref{sec:object_embeddings}, each distribution is represented as a
sample set $\{x_i\}_{i=1}^n$, and the embedder maps this set to a fixed-size
embedding $\embedding$ that parameterizes the decoder. For distributions, the decoder
$\decoder$ is a conditional generative model: it draws samples from the embedded
distribution and evaluates their density. Following the general training procedure,
each sample set is split in half: one half produces the embedding, the other scores
the decoder on held-out samples, and the embedder and decoder are trained jointly.

\begin{figure}[t]
    \centering
    \includegraphics[width=\textwidth]{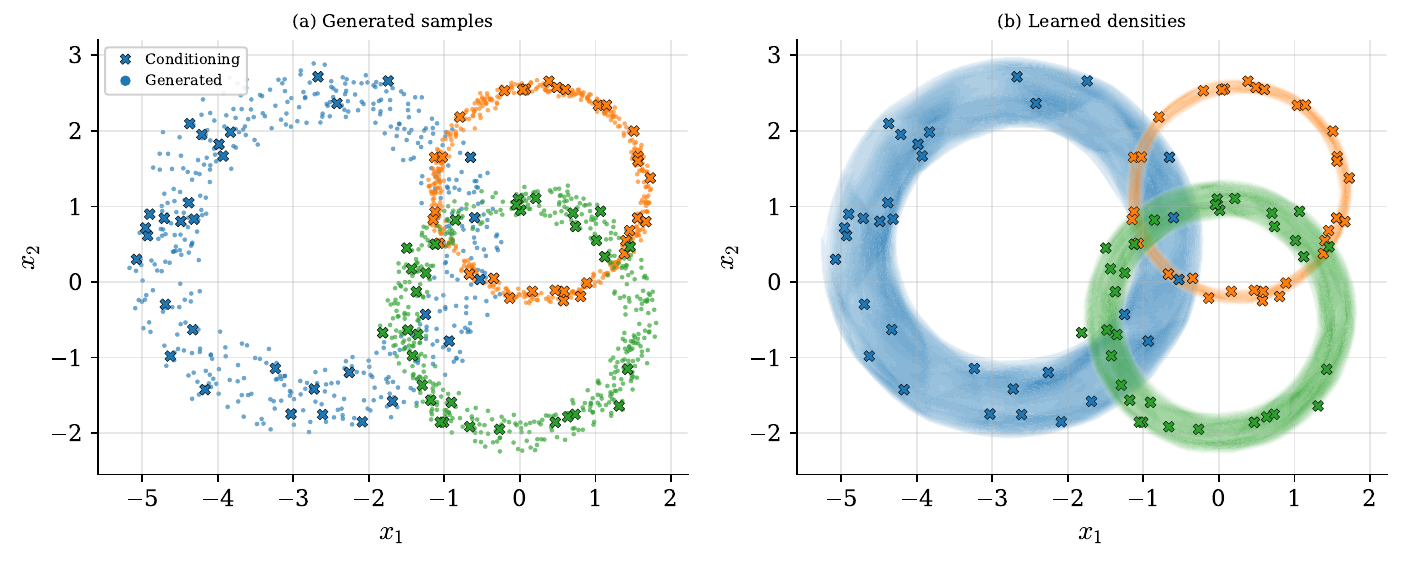}
    \caption{%
        \textbf{Distribution embeddings on the donut family.} \textbf{(a)} Three donuts
        (color-coded), each embedded from a small conditioning set of samples drawn from it
        (crosses); the decoder then generates new samples (dots) from that embedding, recovering
        the donut the conditioning points were drawn from. \textbf{(b)} The density the same
        embedding induces over the plane, evaluated on a grid. From only a handful of conditioning
        samples, the embedding captures each ring's center, radius, and width.%
    }
    \label{fig:distribution_embeddings}
\end{figure}

The embedder is a Set Transformer with induced set attention blocks (ISABs)
\citep{lee2019st}. To help it capture higher moments of the distribution, we extend it
into multiple stages: each stage operates on the original samples combined with a
pooled summary of the previous stage's output. Unlike stacking ISABs, as in the
original architecture, this multi-stage conditioning applies to any set encoder, not
only Set Transformers; we give pseudocode in Appendix~\ref{app:distribution_embeddings}.
The embedding $\embedding$ parameterizes the decoder $\decoder$, which can be any
generative model. Our experiments use both normalizing flows and flow matching. Each is trained
with its usual loss; see Appendix~\ref{app:background} for details on these models. Further
details and hyperparameters are in Appendix~\ref{app:distribution_embeddings}.

We evaluate on the donuts family introduced in Section~\ref{sec:object_embeddings}.
Parameter ranges are in Appendix~\ref{app:distribution_embeddings}.
Figure~\ref{fig:distribution_embeddings} shows three donuts from the family.
From only $32$ samples, the embedding captures each donut well enough for the decoder to
reproduce it both as fresh samples (left) and as a density over the plane (right); both
views trace the true distribution reliably.

Table~\ref{tab:distributions_results} reports reconstruction quality in terms of negative log-likelihood (NLL) and sliced Wasserstein distance (SWD). NLL measures density reconstruction, while SWD measures sample reconstruction. Furthermore, we report the recovery error for each of the three parameters. We compare our approach against two baselines, each fit to every sample set independently: Kernel Density Estimation (KDE), and a parametric fit that recovers the three donut parameters by method of moments and then samples and evaluates density from the fitted donut analytically. With either decoder, our approach outperforms KDE on all five metrics by a large margin, and only trails the parametric fit in radius and width recovery, and even there only marginally. Note that the parametric fit is only applicable because it was given the parametric form of the donut family, which is rarely available in practice; our approach requires no such knowledge and works from samples alone.

The embedding also recovers the family's intrinsic dimensionality. Although the embeddings
live in $\R^{16}$, the donut family has only four scalar degrees of freedom (the two center
coordinates, the radius, and the width), and the first four principal components of the
embeddings explain $99.5\%$ of their variance for the flow-matching head ($97.7\%$ for the
normalizing flow), each aligning almost one-to-one with a single generative property
($|r| \geq 0.95$). Appendix~\ref{app:distribution_embeddings} reports further experiments that
exploit this structure, including embedding arithmetic and interpolation.

\begin{table}[t]
  \centering
  \footnotesize
  \caption{Donut reconstruction at $n{=}256$ embedding samples (mean $\pm$ SEM over 3 seeds, lower is better). FM and NF are our flow matching and normalizing flow decoders on a shared Set Transformer embedder; KDE and a method-of-moments parametric fit are per-sample-set baselines.}
  \label{tab:distributions_results}
  \begin{tabular}{lrrrrr}
    \toprule
    Model & NLL $\downarrow$ & SWD $\downarrow$ & Center $\downarrow$ & Radius $\downarrow$ & Width $\downarrow$ \\
    \midrule
    FM & $1.218 \pm 0.017$ & $0.070 \pm 0.001$ & $0.059 \pm 0.003$ & $0.012 \pm 0.000$ & $0.066 \pm 0.011$ \\
    NF & $1.212 \pm 0.012$ & $0.063 \pm 0.000$ & $0.048 \pm 0.001$ & $0.013 \pm 0.001$ & $0.061 \pm 0.007$ \\
    \midrule
    KDE & $2.011$ & $0.152$ & $0.082$ & $0.048$ & $0.627$ \\
    Parametric fit & $1.332$ & $0.077$ & $0.081$ & $0.010$ & $0.049$ \\
    \bottomrule
  \end{tabular}
\end{table}

\section{Function Embeddings}
\label{sec:function_embeddings}

We now turn to functions. A family of functions is a set that shares a common
structure, such as polynomials of bounded degree, or the solutions of a given
differential equation under varying initial conditions and parameters. A single
sample from a function $f$ is now an input--output pair $(x, f(x))$.

As described in Section~\ref{sec:object_embeddings}, each object 
is represented as a sample set $\{(x_i, f(x_i))\}_{i=1}^n$, and the
embedder maps this set to a fixed-size embedding that parameterizes the decoder.
For functions, the decoder is a pointwise evaluator: given a query point $x$ in
the function's domain, it produces an output $\hat{f}(x)$ approximating $f(x)$.
Following the general training procedure, each sample set is split in half: one
half produces the embedding, the other scores the decoder on held-out samples. 
The embedder and decoder are trained jointly to minimize the error between
$\hat{f}(x)$ and $f(x)$.

In practice, the embedder can be the same multi-stage Set Transformer used in
Section~\ref{sec:distribution_embeddings}, operating on samples formed by
concatenating $x$, $f(x)$ (and optionally the pooled output of the previous
stage). While this already works well, we find it better to project $x$, $f(x)$,
and the pooled output separately into a learned feature space and multiply the three
projections pointwise before passing the resulting set to the Set Transformer.
This three-way product incorporates the pooling information directly, rather than
through concatenation, and yields a stronger embedder; the results we report use
it, and we compare the two in Appendix~\ref{app:function_embeddings}. The decoder
$\decoder$ is a single-head attention layer: it maps the query point $x$ to a query
vector and the embedding $\embedding$ to a set of synthetic key and value vectors,
then attends the query over them to predict $\hat{f}(x)$. This performs better than a
plain feedforward network that concatenates $x$ and $\embedding$.

The loss $\lossfn$ can be any regression loss, such as the mean squared error or the
Huber loss. All our experiments use the relative $L^2$ error,
$\lossfn = \mathbb{E}_{f \sim \family}\!\left[ \lVert f - \hat{f} \rVert_2 / \lVert f \rVert_2
\right]$, as both training loss and evaluation metric, where the norms are
taken over the query points. Normalizing by $\lVert f \rVert_2$ makes the error
insensitive to the scale of the target, and it is standard in operator
learning \citep{lu2021deeponet, li2020fno}. Further details, including
architectures and hyperparameters, are in Appendix~\ref{app:function_embeddings}.

We train and evaluate function embeddings on three families: \emph{polynomials of
bounded degree}, and the solutions of the \emph{damped harmonic oscillator} and
\emph{Lotka--Volterra} ordinary differential equations under varying initial
conditions and parameters. Descriptions and parameter ranges are in
Appendix~\ref{app:function_embeddings}.

We further show that the learned embeddings can serve as inputs and outputs of
\emph{operator networks}, which generalize ordinary functions to maps between
infinite-dimensional function spaces. Because our embeddings represent whole
functions as finite-dimensional vectors, an ordinary neural network becomes a
neural operator: it maps between compact vectors rather than predicting entire
functions, while the decoder expands each result into a function at arbitrary
query points. A single embedding also feeds many heads at once, so the decoder and
every operator share one representation; when this reuse is not needed and
operator learning is the only goal, the reconstruction phase can be skipped and
the operator trained directly. We train two such operators: for polynomials, one
mapping a function's embedding to that of its derivative; for ODEs, one mapping
initial conditions and parameters to the embedding of the solution. Both are
simple feedforward networks; see Appendix~\ref{app:function_embeddings}.

\begin{figure}[t]
    \centering
    \includegraphics[width=\textwidth]{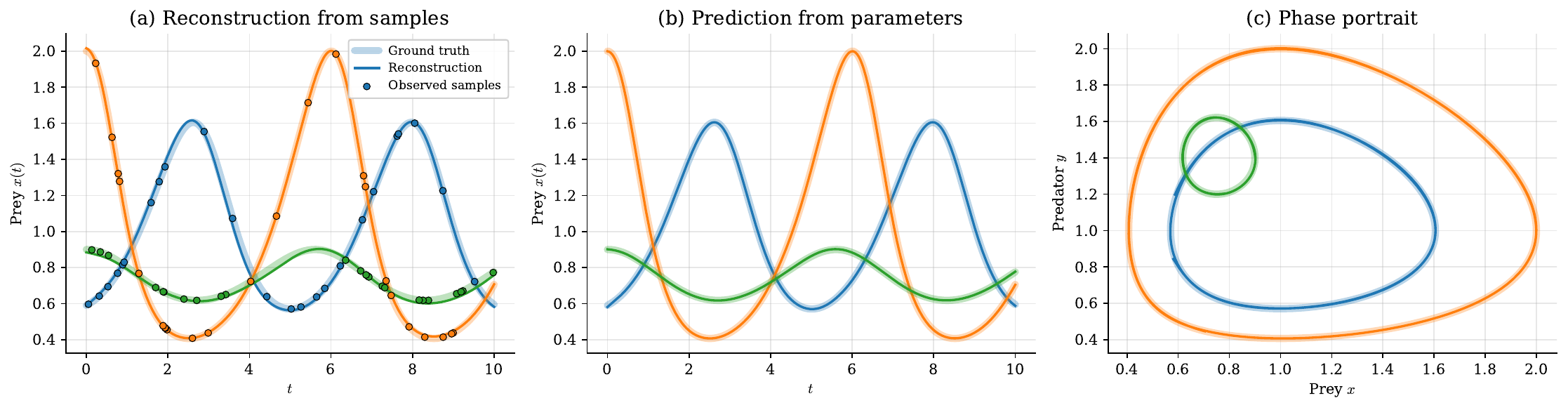}
    \caption{%
        \textbf{Function embeddings on the Lotka--Volterra family.} \textbf{(a)}
        Reconstruction of three solutions (color-coded) from observed samples: the
        embedder maps the samples (dots) to an embedding from which the decoder
        reconstructs the prey trajectory $x(t)$ (solid line, ground truth shaded).
        \textbf{(b)} Prediction of the same solutions from their parameters
        $(x_0, y_0, \alpha, \beta, \gamma, \delta)$ alone, with no samples, via the
        operator. \textbf{(c)} The predicted trajectories drawn as a predator--prey
        phase portrait. The same decoder recovers the trajectory whether the embedding
        is formed from samples or predicted from parameters.%
    }
    \label{fig:function_embeddings}
\end{figure}

Figure~\ref{fig:function_embeddings} shows both reconstruction and operator
prediction on the Lotka--Volterra family. From only a small sample set, the
embedding captures enough structure for the decoder to reconstruct the full
trajectory (left), while an operator trained to output an embedding can predict the
solution from its initial conditions and parameters alone (middle and right).

Table~\ref{tab:fn_families} reports both tasks across the three families, and
Table~\ref{tab:fn_scaling} reports performance on a generalized $n$-dimensional
damped harmonic oscillator as the state dimension grows. The families cover
qualitatively different behaviors: the damped-oscillator and Lotka--Volterra
solutions are oscillatory across a wide range of amplitudes, the former decaying to
zero and the latter periodic, while the polynomials range from near-zero constants
and linear functions to quadratics and cubics with values in the tens and hundreds.
In each case, a single model captures its family's full range of behaviors,
and the approach scales to functions of up to $32$ dimensions.

The two operators behave differently relative to the sample-inferred embedding
(the \emph{Embedding} column). The ODE solution operator maps clean, fixed
parameters to an embedding and can therefore be \emph{more} accurate than an
embedding inferred from a noisy, finite sample set. The differential operator
instead takes such a sample-inferred embedding as input, inheriting its noise and
adding its own, so its error exceeds the embedding's.

\begin{table}[t]
    \centering
    \small
    \caption{%
        Relative $L^2$ reconstruction error by family (mean $\pm$ SEM over 3
        seeds). \emph{Embedding}: the function is reconstructed from an embedding inferred
        from a sample set; \emph{Operator}: from an embedding produced by an operator. For
        polynomials, operator error is reported on the output derivative and indexed by its
        degree. A degree-three derivative is omitted, as it would require a degree-four input.%
    }
    \label{tab:fn_families}
    \begin{tabular}{@{}l cc@{}}
        \toprule
        Family & Embedding & Operator \\
        \midrule
        Polynomials & $0.0143 \pm 0.0011$ & $0.0185 \pm 0.0022$ \\
        \quad Degree 0 & $0.0016 \pm 0.0001$ & $0.0133 \pm 0.0013$ \\
        \quad Degree 1 & $0.0103 \pm 0.0004$ & $0.0151 \pm 0.0014$ \\
        \quad Degree 2 & $0.0201 \pm 0.0019$ & $0.0271 \pm 0.0039$ \\
        \quad Degree 3 & $0.0251 \pm 0.0024$ & \textemdash \\
        Damped Oscillator & $0.0173 \pm 0.0029$ & $0.0056 \pm 0.0002$ \\
        Lotka--Volterra & $0.0047 \pm 0.0001$ & $0.0020 \pm 0.0000$ \\
        \bottomrule
    \end{tabular}
\end{table}

\begin{table}[t]
    \centering
    \small
    \caption{%
        Damped-oscillator dimension scaling: relative $L^2$ reconstruction error
        vs.\ state dimension (mean $\pm$ SEM over 3 seeds; columns as in
        Table~\ref{tab:fn_families}).%
    }
    \label{tab:fn_scaling}
    \begin{tabular}{@{}r cc@{}}
        \toprule
        Dim. & Embedding & Operator \\
        \midrule
        $2$ & $0.0173 \pm 0.0029$ & $0.0056 \pm 0.0002$ \\
        $4$ & $0.0178 \pm 0.0010$ & $0.0070 \pm 0.0006$ \\
        $8$ & $0.0279 \pm 0.0003$ & $0.0122 \pm 0.0006$ \\
        $16$ & $0.0404 \pm 0.0042$ & $0.0199 \pm 0.0013$ \\
        $32$ & $0.0645 \pm 0.0015$ & $0.0348 \pm 0.0015$ \\
        \bottomrule
    \end{tabular}
\end{table}

\section{Graph Embeddings}
\label{sec:graph_embeddings}

Finally, we instantiate our framework on families of graphs. We consider
undirected labeled graphs $\mathcal{G} = (V, E)$ with no self-loops, where $V$
is the vertex set and $E$ the edge set. A family is a collection of graphs that
share a common vertex set $V$ while the edge set $E$ varies. A single sample is
now an edge: an unordered pair of vertices.

Following the procedure laid out in Section~\ref{sec:object_embeddings}, a graph is
represented solely as a set of randomly sampled edges $\{e_1, \dots, e_n\}$. Each
edge is encoded as a two-hot vector, with ones at the positions of its two
vertices, so that the embedder is invariant to their order. The embedder maps this
set to a fixed-size embedding and uses the same multi-stage Set Transformer
architecture as in the previous sections. The embedding parameterizes a decoder
whose goal is to recover a \emph{Laplacian representation} of the vertices: a
spectral representation of a graph that is a useful state representation for reinforcement learning
\citep{gomez2024allo}. Following the general training procedure, the sampled edges
are split in half, one half producing the embedding and the other scoring the
decoder, and the two networks are trained jointly.

The decoder is trained with the ALLO objective (Appendix~\ref{subsec:allo}). In
ALLO both the vertex-representation network and the lower-triangular matrix of
dual variables are fit to a single graph. We make both depend on the embedding
instead: the decoder maps each vertex to its representation vector conditioned on
$\embedding$, and we exchange the fixed matrix of dual variables for a separate
neural network that predicts its entries from $\embedding$. A single model thus
produces the Laplacian representation of any graph in the family. Further details
are in Appendix~\ref{app:graph_embeddings}.

We train and evaluate on three families: unweighted random trees, scale-free
networks, and weighted gridworlds. Detailed descriptions and parameter ranges are
in Appendix~\ref{app:graph_embeddings}.
The quality of learned representation and embeddings is assessed in two ways. First, by its \emph{geometry}:
whether the distance between vertices effectively separates edges from non-edges. This is
meaningful because the Euclidean distance between two vertices in the Laplacian
representation is known to correlate with their temporal distance in the graph
\citep{gomez2024allo}. Second, by \emph{reconstruction}: whether a learned thresholding strategy can
recover adjacency, predicting an edge between two vertices when their
representation distance falls below a per-vertex threshold. We report two
thresholding strategies. The \emph{vertex-degrees} strategy is an oracle that uses
each vertex's true degree to set its threshold, whereas the \emph{medians network}
is trained, solely from the sampled edges, to predict the median distance from a
vertex to its neighbors. It minimizes the mean absolute error between the
representation distances of sampled edges and the predicted medians:
\begin{equation*}
\lossfn_{\text{med}} = \mathbb{E}_{(u,v) \in E}\Big[\,
  \big| d(u, v) - m_u \big| + \big| d(u,v) - m_v \big| \,\Big],
\end{equation*}
where $d(u, v) = \lVert h_u - h_v \rVert_2$ is the distance between the
representation vectors of $u$ and $v$, and $m_u$ is the median distance
the network predicts for vertex $u$ to its neighbors.

\begin{figure}[t]
    \centering
    \includegraphics[width=\textwidth]{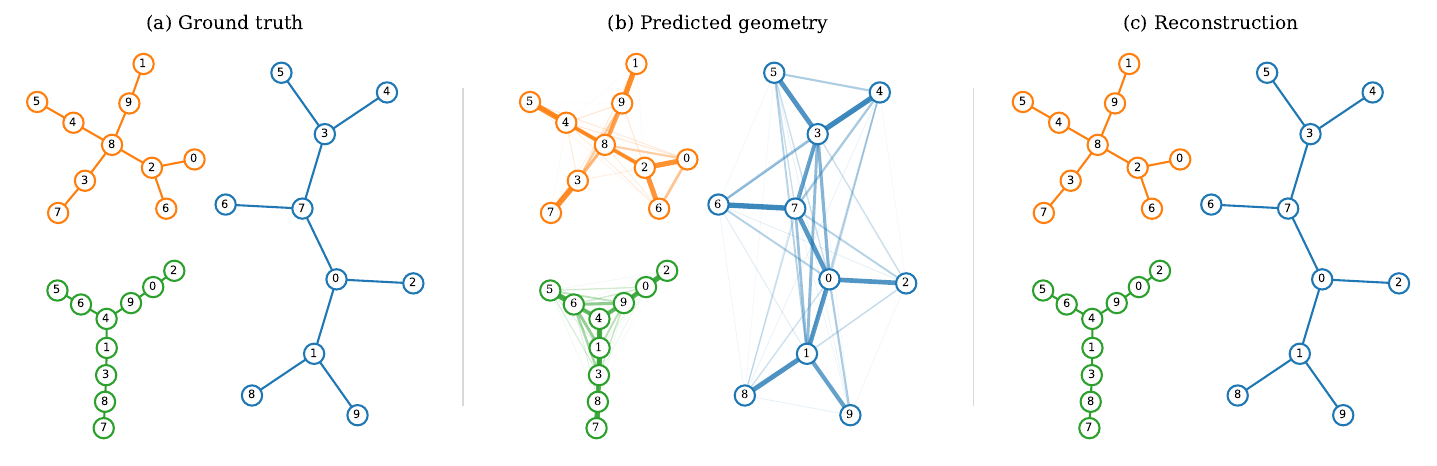}
    \caption{%
        \textbf{Graph embeddings on the random-trees family.} For three sampled trees
        (color-coded), we show \textbf{(a)} the ground-truth graphs, \textbf{(b)} the
        predicted geometry (each pair of vertices joined by a line that grows thicker and
        darker as their representation distance shrinks), and \textbf{(c)} the
        reconstruction from thresholding those distances with the medians network. The
        tree structure is already visible in the geometry alone.%
    }
    \label{fig:graph_embeddings}
\end{figure}

Figure~\ref{fig:graph_embeddings} shows three randomly sampled trees, together with
their predicted geometry and reconstructions. The geometry panel draws, for each
pair of vertices, a line whose thickness and darkness grow as their representation
distance shrinks. From the predicted geometry alone, the structure of each tree is
clearly visible. To turn these distances into binary edge/non-edge decisions, we
apply per-vertex thresholds from the trained medians network; the rightmost panel
shows the resulting reconstruction.
Table~\ref{tab:graph_families} reports the geometry ROC-AUC and the reconstruction
F1 of the degree oracle and the medians network, for random trees and scale-free
graphs at $n = 10$ and $n = 20$ vertices. Additional results, including the
weighted gridworld family, are in Appendix~\ref{app:graph_embeddings}.

The ROC-AUC scores show that the model separates edges from non-edges well across
all four settings, and on the $n = 10$ families the medians network nearly matches
the degree oracle. On the larger $n = 20$ families, both reconstruction strategies
degrade, largely because the geometry is learned less accurately there, as
reflected in the lower ROC-AUC. The medians network predicts an edge between $u$
and $v$ whenever their distance falls below $m_u + m_v$, the sum of the two
endpoints' medians; since each median is itself roughly a typical edge distance,
this is a loose rule that tends to over-predict edges. It also learns these medians
from only the sampled edges, the same data the embedder and decoder see, so giving
it access to labeled data raises its F1; we report additional
reconstruction-head numbers in Appendix~\ref{app:graph_embeddings}. Scale-free
graphs are harder than trees at both sizes, most likely because they are denser and
contain hubs: a hub has many neighbors at a wide range of distances, so a single
median cannot separate its edges from non-edges cleanly.

\begin{table}[t]
    \centering
    \small
    \caption{%
        Geometry and reconstruction quality by family (mean $\pm$ SEM over 3 seeds).
        Geometry ROC-AUC is the threshold-free edge/non-edge separation; the two F1
        columns are degree-oracle and learned-head (medians) reconstruction.%
    }
    \label{tab:graph_families}
    \begin{tabular}{@{}l ccc@{}}
        \toprule
        Family & ROC-AUC & Vtx-degree F1 & Medians F1 \\
        \midrule
        Random trees ($n=10$) & $0.993 \pm 0.001$ & $0.983 \pm 0.001$ & $0.922 \pm 0.005$ \\
        Random trees ($n=20$) & $0.953 \pm 0.006$ & $0.781 \pm 0.028$ & $0.449 \pm 0.064$ \\
        Scale-free ($n=10$) & $0.980 \pm 0.001$ & $0.958 \pm 0.002$ & $0.793 \pm 0.002$ \\
        Scale-free ($n=20$) & $0.847 \pm 0.010$ & $0.613 \pm 0.022$ & $0.331 \pm 0.008$ \\
        \bottomrule
    \end{tabular}
\end{table}

\section{Discussion}
\label{sec:discussion}

We demonstrated the generality and usefulness of our framework on three very distinct
higher-order objects: distributions, functions, and graphs. Appendix~\ref{sec:related_work}
underscores how distinct these objects are, as each has historically been approached with
its own domain- and object-specific methods.

\paragraph{Distributions.} Distribution embeddings, as presented in Section~\ref{sec:distribution_embeddings}, have already been applied to state tracking in partially observable domains as part of Neural Bayesian Filtering \citep{solinas2026nbf}, which in turn underpins an online POMDP solver, Belief Embedding Tree Search (BETS) \citep{solinas2026bets}. Because the belief state is now a single fixed-size vector, policies and value functions can be learned directly on the embedding, which was not possible with the particle-based representations they replace. Compared to the closest prior work, our framework generalizes the Neural Statistician \citep{edwards2016statistician} by admitting any decoder, not just a variational autoencoder, and it goes beyond the continuous distributions to which Generative Distribution Embeddings \citep{fishman2026gde} have so far been applied.
BETS embeds discrete distributions by supplying a discrete generative model as the decoder, and we discuss discrete extensions of normalizing flows and flow matching in Appendix~\ref{app:background}.

\paragraph{Functions.} Similar to distribution embeddings, function embeddings reconstruct functions accurately from a handful of samples and even scale to higher-dimensional problems. They also demonstrate how useful a fixed-size vector representation of a higher-order object can be: unlike specialized approaches such as DeepONet \citep{lu2021deeponet} and FNO \citep{li2020fno}, our embeddings let ordinary neural networks act as neural operators. Because these operators act on the learned embeddings rather than on raw function values, they carry none of the rigid input assumptions, such as regular grids, fixed sensor locations, or fixed resolutions, that these methods impose, yet they can still exploit that structure when it is available. In our experiments a single model maps parameters directly to ODE solutions, and solving a PDE requires nothing new: it reduces to mapping one embedding to another, the same operation we demonstrated with the differential operator that sends a function's embedding to that of its derivative. Finally, because Physics-Informed Neural Networks (PINNs) \citep{raissi2019pinns} only alter the training loss, our framework is open to them as well.

\paragraph{Graphs.} In Section~\ref{sec:graph_embeddings} we took ALLO, an existing single-object model, and lifted it to a whole family for the first time. As originally proposed by \citet{gomez2024allo}, it operates on one graph at a time, so a new graph requires retraining from scratch. Using it as the object-specific decoder $\decoder$, we made one small change: the fixed lower-triangular matrix of dual variables became a neural network that predicts its entries from the embedding. The lifted model learns vertex representations across a whole family of graphs, recovering a geometry that separates edges from non-edges well. Recovering the exact adjacency is harder, for two reasons: ALLO optimizes for spectral geometry and not reconstruction; and thresholding distances against a single per-vertex median cannot cleanly separate edges from non-edges at every vertex, particularly at high-degree hubs. Reconstruction-oriented approaches, such as those discussed in Appendix~\ref{sec:related_work}, may be better suited to that task. Finally, because the embedding summarizes a graph as a single vector, it is a natural conditioning signal for reinforcement learning: an agent could act in an unknown environment, gather enough transitions to embed the local structure of its underlying Markov Decision Process, and condition its policy on that embedding.

\section{Conclusion}
\label{sec:conclusion}

We introduced the first framework general enough to learn compact
representations of higher-order objects and to generalize across families of
them. The framework accommodates any higher-order object through three
user-provided components: an object sampler, a parameterizable fitting model,
and a training loss. We instantiated it on three objects that have each
historically demanded vastly different architectures and training procedures:
distributions, functions, and graphs. In every case, a single trained model
represents an entire family and generalizes to members unseen during training.
Specific applications of this approach (notably distribution embeddings) have already been shown to be useful for tracking beliefs in POMDPs and in building cell dynamics models.

\subsection*{Acknowledgments}

This work was supported in part by the Canada CIFAR AI Chairs program and the Alberta Machine Intelligence Institute (Amii) and the Natural Sciences and Engineering Research Council of Canada (NSERC), and was enabled in part by support provided by Prairies DRI and the Digital Research Alliance of Canada (\url{https://www.alliancecan.ca}).
R. Haluška and M. Schmid were supported by the grant no. 25-18031S of the Czech Science Foundation (GAČR) and the Charles University Grant Agency (GAUK) project no. 326525. Additional computational resources were provided by the e-INFRA CZ project (ID:90254), supported by the Ministry of Education, Youth and Sports of the Czech Republic.

\subsection*{AI Use Statement}

Large language models (LLMs) assisted us with writing and coding. They were used to refine writing, discover relevant prior work, and implement and test the experiment pipeline. We revised and verified all LLM output carefully and take responsibility for the final content of this work, including any text, claims, or artifacts produced with the aid of generative AI.

\subsection*{Reproducibility Statement}

The source code implementing the proposed method and the experiments presented in the paper has been made available as supplementary material and will be open-sourced on GitHub upon publication. The appendices provide further implementation details, such as hyperparameter ranges.

\bibliography{refs.bib}
\bibliographystyle{iclr2027_conference}

\newpage
\appendix
\section{Background}
\label{app:background}

\subsection{Generative Modeling}

\textbf{Normalizing flows} \citep{papamakarios2021nfs} model a
distribution with an invertible, differentiable map $f_\psi : \reals^d
\to \reals^d$ that pushes a tractable base $\prior$ (a standard Gaussian) to
the target. A sample is $\state = f_\psi(z)$ with $z \sim \prior$, and
its density follows from the change-of-variables formula
\[
  \log p_\psi(\state)
  = \log \prior\!\left(f_\psi^{-1}(\state)\right)
  + \log \left| \det \frac{\partial f_\psi^{-1}}{\partial \state}(\state) \right|.
\]
Architectures such as RealNVP \citep{dinh2016realnvp} and neural spline flows
\citep{durkan2019splineflows} keep $f_\psi^{-1}$ and this log-determinant
tractable by composing layers with triangular Jacobians, and are trained by
maximizing the log-likelihood of the data. Flows live on continuous spaces; a
discrete support is reached through \emph{variational dequantization}
\citep{ho2019vardeq}, which adds learned noise to each point so the categorical
target becomes a continuous density, at the price of an implicit ordinal
structure over classes that may in fact be unordered.

\textbf{Flow matching} \citep{lipman2022fm} is a continuous-time analogue
that replaces the composition of invertible layers with a learned velocity
field $v_\psi(z, \tau)$ over $\tau \in [0, 1]$. Sampling integrates
$\dot{z}_\tau = v_\psi(z_\tau, \tau)$ from $z_0 \sim \prior$ and returns
$\state \defeq z_1$, and densities follow from integrating the instantaneous
change of variables along the same trajectory. Given the optimal-transport
interpolant $z_\tau = (1-\tau)\, z + \tau\, \state$ between $z \sim \prior$ and
a target sample $\state \sim p$, training regresses onto the target velocity,
\[
  \mathcal{L}_{\text{FM}}(\psi)
  = \E_{\tau, z, \state}\!\left\| v_\psi(z_\tau, \tau) - (\state - z) \right\|^2 .
\]
Dropping the closed-form-inverse requirement makes the transport more
expressive, at the cost of an ODE solve for both sampling and density. Its
categorical analogue, \emph{discrete flow matching} \citep{gat2024dfm},
interpolates between the data and a factorized noise distribution over classes
and trains a denoiser to recover the clean configuration, which is integrated
in reverse to sample; unlike dequantization it acts on the discrete support
directly, with no ordinal bias.

\subsection{Augmented Lagrangian Laplacian Objective}
\label{subsec:allo}

An undirected graph without self-loops is fully described by its \emph{graph
Laplacian} $\mathbf{L} = \mathbf{D} - \mathbf{A}$, where $\mathbf{D}$ is the
diagonal degree matrix and $\mathbf{A}$ the (weighted) adjacency matrix. The
\emph{Laplacian representation} assigns each vertex $i$ the $d$-dimensional
vector formed from the $i$-th component of the $d$ smallest eigenvectors of
$\mathbf{L}$. Distances in this representation encode temporal structure of the
underlying state space, making it a useful state representation for
reinforcement learning \citep{gomez2024allo}.

The augmented Lagrangian Laplacian objective (ALLO) \citep{gomez2024allo}
learns this representation for a single graph from its sampled edges. It is a
saddle-point objective that recovers the true eigenvectors and eigenvalues of
$\mathbf{L}$ with little hyperparameter tuning, jointly optimizing a network
that maps each vertex to its representation vector against an explicit
lower-triangular matrix of dual variables $(\beta_{jk})_{1 \leq k \leq j \leq
d}$, one per pair of eigenvectors. The diagonal entries $\beta_{jj}$ converge
to values proportional to the eigenvalues, and the off-diagonal entries
$\beta_{jk}$ enforce orthogonality between the eigenvectors.

\section{Related Work}
\label{sec:related_work}

To the best of our knowledge, no prior approach represents and learns from
families of general higher-order objects. Existing work targets specific object
types: mostly functions, through operator learning, and more recently
distributions across a range of domains.

\paragraph{Distributions.} Kernel mean embeddings (KMEs) \citep{muandet2017kme}
are a classical non-parametric way to represent a distribution as a point in a
reproducing kernel Hilbert space, but they scale poorly with dimension, are not
natively generative, and require a kernel whose choice and hyperparameters
usually demand domain knowledge. The Neural Statistician
\citep{edwards2016statistician} instead learns distribution embeddings from
i.i.d.\ samples and uses them to condition a variational autoencoder (VAE).
Generative Distribution Embeddings (GDEs) \citep{fishman2026gde} generalize both by
allowing an arbitrary conditional generative model as the decoder; they also
justify distributional invariance theoretically and show that latent distances
correlate with the Wasserstein-2 distance, though they have been applied only
to continuous distributions in computational biology. At the same time,
\citet{solinas2026nbf} independently proposed distribution embeddings to
represent belief states in a filtering algorithm for partially observable state
tracking, outperforming traditional filters.

\paragraph{Functions.} Neural Processes (NPs)
\citep{garnelo2018neuralprocesses} combine Gaussian processes and neural
networks to learn distributions over functions: they encode a set of
input-output pairs into a distribution over latent codes, which a decoder
evaluates at a query input to predict the function's value there, and are
trained by maximizing an ELBO. DeepONet \citep{lu2021deeponet} and FNO
\citep{li2020fno} are neural operators that learn maps between function spaces,
typically PDE solution operators, and so act on families of related functions;
DeepONet uses branch and trunk networks, while FNO learns the operator in the
spectral domain via Fourier transforms. Both carry function-specific biases:
DeepONet needs conditioning points on a predefined grid and FNO assumes uniform
grids. Later variants such as GeoFNO \citep{li2023geofno} relax some of these
but remain tied to functions (see \citet{kovachki2023neuraloperator} for an
overview). Finally, function encoders \citep{ingebrand2024functionencoders}
represent a function as a linear combination of learned basis functions, first
for reinforcement learning and later for transfer learning
\citep{ingebrand2025functiontransfer}.

\paragraph{Graphs.} Most graph representation learning embeds \emph{nodes} rather than
whole graphs: GraphSAGE \citep{hamilton2017graphsage}, for example, learns an inductive
node encoder that generalizes to unseen nodes and graphs, but yields one vector per node
rather than a single representation of the graph. graph2vec \citep{narayanan2017graph2vec}
instead embeds an entire graph as a fixed-size vector, learned unsupervised by treating
rooted subgraphs as words in the manner of document embeddings, though the resulting
representation is task-agnostic and tied to graph-specific substructures. A separate line
of work learns to \emph{generate} graphs: GraphVAE \citep{simonovsky2018graphvae} decodes a
graph from a latent code with a variational autoencoder, GraphRNN \citep{you2018graphrnn}
builds graphs autoregressively, and DiGress \citep{vignac2022digress} does so through
discrete denoising diffusion. These models capture a distribution over graphs but do not
expose a reusable embedding of any given graph.

Compared to the approaches above, our framework is not tied to a single object type and
bakes in no domain-specific assumptions, though such biases can still be reintroduced
through the decoder $\decoder$ and the loss $\lossfn$ when useful. This generality is what
lets many of these methods be recast as special cases of our framework under particular
choices of $\decoder$ and $\lossfn$, and it lets any existing single-object model serve as
a decoder, making the framework straightforward to adapt to new problems and domains.

\section{Multi-Stage Set Encoder}
\label{app:multistage_embedder}

The embedder wraps any permutation-equivariant set encoder in a multi-stage loop. Stage $k$ re-reads
the \emph{original} samples, each concatenated with a pooled summary of the previous stage's output,
so that later stages can compute features relative to the set-level context, such as the spread of the
samples around their estimated mean (a second moment). Unlike stacking induced set attention blocks,
which is specific to the Set Transformer, this wrapper applies to any set encoder, including DeepSets
and the Set Transformer we use. Algorithm~\ref{alg:multistage} states the procedure: each
$\mathrm{SetEnc}_k$ is a set encoder mapping $N$ inputs to $N$ features, and $\mathrm{Pool}$ is a
weighted mean (our default) or an attention pooling.

\begin{algorithm}
\caption{Multi-stage set encoder}
\label{alg:multistage}
\begin{algorithmic}[1]
\Require samples $\{x_i\}_{i=1}^N$ with weights $\{w_i\}_{i=1}^N$; stages $\mathrm{SetEnc}_{1:K}$; pooling $\mathrm{Pool}$
\Ensure embedding $z$
\State $\bar{w}_i \gets w_i \,/\, \textstyle\sum_j w_j$ \Comment{normalize the sample weights}
\State $c \gets \varnothing$ \Comment{no context for the first stage}
\For{$k = 1$ \textbf{to} $K$}
    \State $h_i \gets x_i$ \textbf{if} $k = 1$ \textbf{else} $[\,x_i \,;\, c\,]$, \quad for all $i$ \Comment{concatenate the running context to each sample}
    \State $(g_1, \dots, g_N) \gets \mathrm{SetEnc}_k(h_1, \dots, h_N;\, w)$ \Comment{permutation equivariant, $N \to N$}
    \State $c \gets \mathrm{Pool}(g_1, \dots, g_N;\, \bar{w})$ \Comment{$N \to 1$; default $c = \textstyle\sum_i \bar{w}_i\, g_i$}
\EndFor
\State \Return $z \gets c$
\end{algorithmic}
\end{algorithm}

\section{Distribution Embeddings}
\label{app:distribution_embeddings}

\subsection{Distribution Families}
\label{app:distribution_families}

\paragraph{Donuts.} Two-dimensional ring-shaped distributions, each fully described
by a center $c \in \R^2$, a radius $r$, and a width $w$. A sample is drawn as
$x = c + r\,a\,(1 + w\,b)$, where $a$ is a uniformly random unit vector and
$b \sim \mathcal{U}[-\tfrac{1}{2}, \tfrac{1}{2}]$, so the points fill an annulus of
mean radius $r$ whose thickness is $r w$. The family
draws the center from $\mathcal{N}(0, I)$, the radius uniformly from $[0.5, 2]$, and
the width uniformly from $[0.01, 0.75]$, spanning both thin and thick donuts.

\subsection{Embedding Properties}
\label{app:distribution_properties}

The main text shows that the embedding space is low-dimensional and aligned with the
donut's generative parameters. Two further experiments test whether it is also
\emph{linear}: whether straight lines and vector arithmetic between embeddings
correspond to meaningful operations on the distributions. Both use the same flow-matching
model from Section~\ref{sec:distribution_embeddings} with $512$-sample embeddings, and every panel below shows the learned density
decoded from a single embedding.

\begin{figure}
    \centering
    \includegraphics[width=1\linewidth]{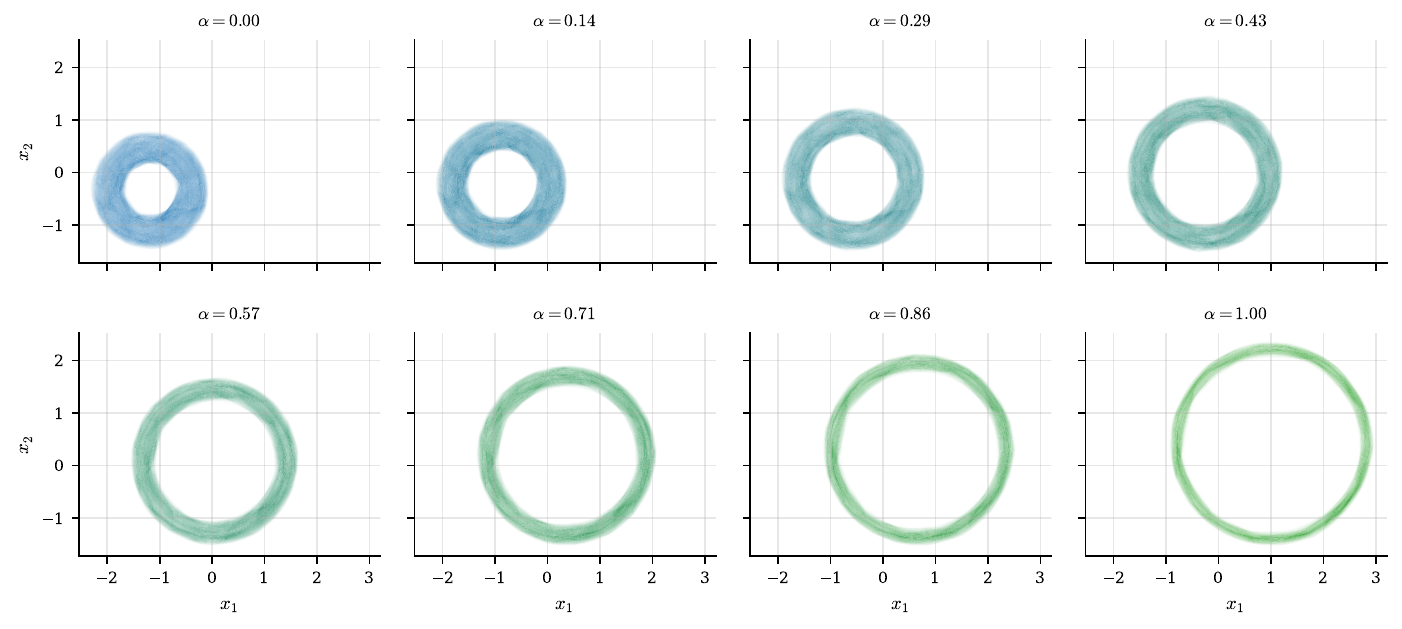}
    \caption{Densities decoded along a straight line between the embeddings of two
    donuts ($\alpha = 0$ and $\alpha = 1$). The interior points are interpolated in
    embedding space, not embedded from samples. Color tracks $\alpha$.}
    \label{fig:donut-interpolation}
\end{figure}

\paragraph{Interpolation.} Figure~\ref{fig:donut-interpolation} embeds two donuts that
differ in center, radius, and width, and decodes the densities along the segment
$z_\alpha = (1 - \alpha)\, z_0 + \alpha\, z_1$ between their embeddings. During
training, every embedding the decoder sees is produced by the encoder from the samples of
some donut; the interior points $z_\alpha$ are not, being mere convex combinations of two
such embeddings, so nothing forces the decoder to treat them sensibly. Yet each decodes to
a single well-formed ring, and the rings translate, grow, and thin as $\alpha$ increases.
This is neither of the two naive interpolations. Mixing the two densities,
$(1 - \alpha)\, p_0 + \alpha\, p_1$, cross-fades one ring into the other and never
produces an intermediate donut, and moving paired samples along straight lines smears the
rings into a cloud that is no donut at all. The embedding instead moves through the
family: every intermediate point is a donut with intermediate center, radius, and width.

\begin{figure}
    \centering
    \includegraphics[width=1\linewidth]{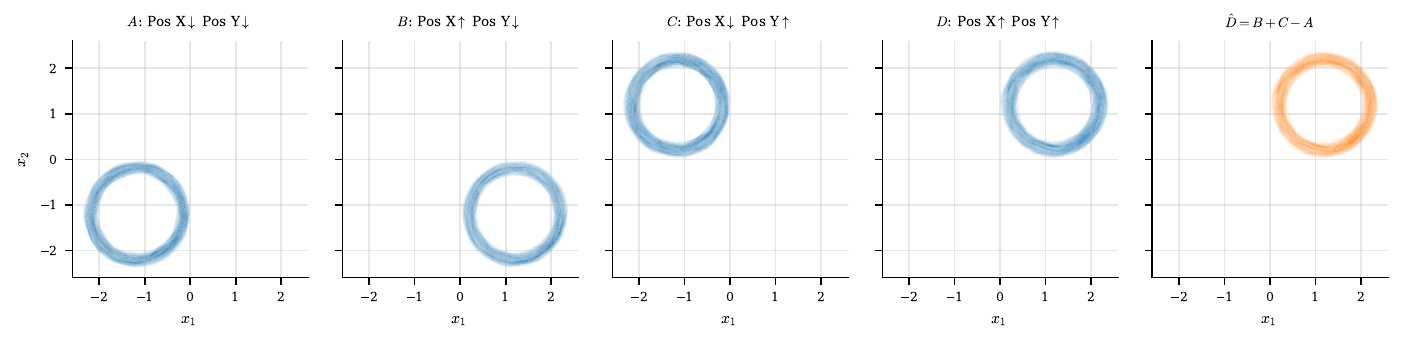}
    \caption{Embedding arithmetic across horizontal and vertical position. Blue panels
    are embedded from samples; the orange panel decodes the prediction
    $\hat{D} = z_B + z_C - z_A$, which matches $D$ to within the sampling-noise floor.}
    \label{fig:donut-composition-position}
\end{figure}

\begin{figure}
    \centering
    \includegraphics[width=1\linewidth]{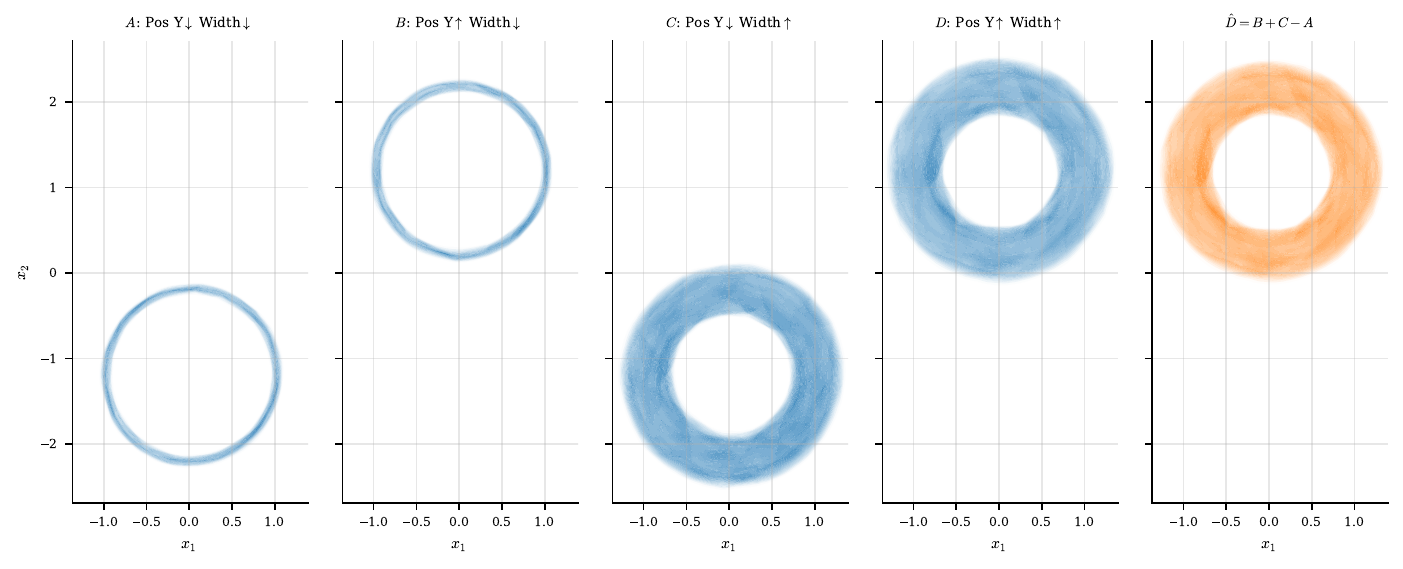}
    \caption{The same identity across vertical position and width. The prediction
    (orange) again matches $D$ to within the noise floor.}
    \label{fig:donut-composition-width}
\end{figure}

\begin{figure}
    \centering
    \includegraphics[width=1\linewidth]{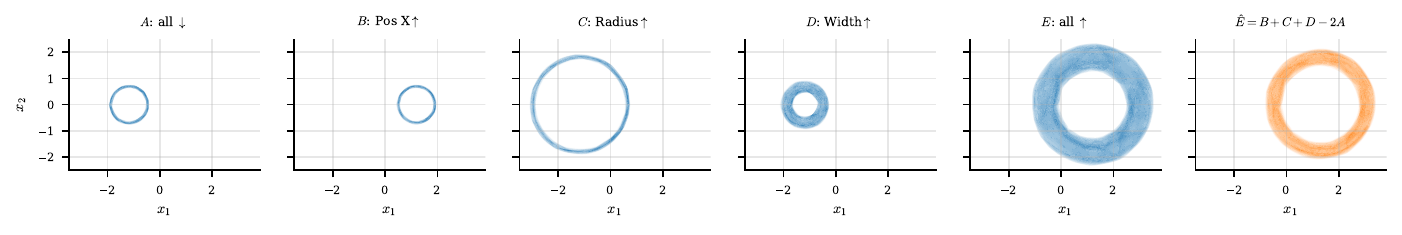}
    \caption{Where additivity breaks: raising horizontal position, radius, and width
    together. The prediction $\hat{E} = z_B + z_C + z_D - 2 z_A$ (orange) is visibly
    thinner than $E$, because a donut's thickness is the product of radius and width.}
    \label{fig:donut-composition-triple}
\end{figure}

\paragraph{Embedding arithmetic.} Starting from a base donut $A$, we raise one
property to obtain $B$ or $C$, and both to obtain $D$. If the properties are encoded
along independent directions, the parallelogram identity $z_D \approx z_B + z_C - z_A$
holds, and decoding its right-hand side predicts $D$ without ever seeing samples of
it. Figures~\ref{fig:donut-composition-position} and~\ref{fig:donut-composition-width}
show this for two positional properties and for a positional and a shape property.
The predictions agree with the true corner to a cosine similarity above $0.999$, with
relative errors of $3.1\%$ and $4.2\%$ of $\|z_D\|$, at the $3.9\%$ floor set by
sampling noise in a $512$-sample embedding. The identity extends to three properties as
$z_E \approx z_B + z_C + z_D - 2 z_A$, and it breaks only where the family itself is
not additive: a donut's absolute thickness is the product of its radius and its relative
width, so raising both interacts multiplicatively, and no additive code can represent a
product. The triple in Figure~\ref{fig:donut-composition-triple}, which raises radius
and width together, misses by $24\%$ and decodes to a visibly thinner ring, whereas
triples avoiding that pair stay near their $5.6\%$ noise floor.

\subsection{Additional Experiments}
\label{app:distribution_additional}

Table~\ref{tab:distributions_results} in the main text reports the operating point
$n = 256$. Tables~\ref{tab:donut-recon} and~\ref{tab:donut-errors} sweep the number of
samples the embedding is computed from, $n \in \{16, 32, 64, 128, 256, 1024\}$, for both
generator heads on the same $1{,}024$ held-out donuts. Reconstruction quality improves
steeply from $n = 16$ to $n = 64$ and saturates by $n = 128$: with more samples the
embedding no longer changes what the decoder produces, so the remaining error reflects
the capacity of the amortized model rather than the size of the sample set. The two heads
tie on NLL, while the normalizing flow is slightly better on SWD at every $n$. The center
and radius errors keep decreasing with $n$ and level off by $n = 128$, whereas the width
error, the noisiest of the three across seeds, stops improving after $n = 32$.

\begin{table}[t]
  \centering
  \footnotesize
  \setlength{\tabcolsep}{4pt}
  \caption{Donut reconstruction quality: held-out NLL and sliced-Wasserstein distance (mean $\pm$ SEM over seeds, lower is better) versus embedding sample count $n$.}
  \label{tab:donut-recon}
  \begin{tabular}{lrrrr}
    \toprule
    & \multicolumn{2}{c}{NLL} & \multicolumn{2}{c}{SWD} \\
    \cmidrule(lr){2-3}\cmidrule(lr){4-5}
    $n$ & FM & NF & FM & NF \\
    \midrule
    $16$ & $1.819 \pm 0.202$ & $1.939 \pm 0.108$ & $0.098 \pm 0.001$ & $0.091 \pm 0.002$ \\
    $32$ & $1.311 \pm 0.028$ & $1.319 \pm 0.013$ & $0.079 \pm 0.001$ & $0.073 \pm 0.000$ \\
    $64$ & $1.223 \pm 0.009$ & $1.215 \pm 0.006$ & $0.072 \pm 0.000$ & $0.066 \pm 0.000$ \\
    $128$ & $1.214 \pm 0.015$ & $1.205 \pm 0.010$ & $0.070 \pm 0.001$ & $0.064 \pm 0.000$ \\
    $256$ & $1.218 \pm 0.017$ & $1.212 \pm 0.012$ & $0.070 \pm 0.001$ & $0.063 \pm 0.000$ \\
    $1024$ & $1.224 \pm 0.017$ & $1.224 \pm 0.014$ & $0.069 \pm 0.001$ & $0.063 \pm 0.001$ \\
    \bottomrule
  \end{tabular}
\end{table}

\begin{table}[t]
  \centering
  \footnotesize
  \setlength{\tabcolsep}{4pt}
  \caption{Donut parameter-recovery error: center, radius, and width of the reconstructed ring (mean $\pm$ SEM over seeds, lower is better) versus embedding sample count $n$.}
  \label{tab:donut-errors}
  \begin{tabular}{lrrrrrr}
    \toprule
    & \multicolumn{2}{c}{Center} & \multicolumn{2}{c}{Radius} & \multicolumn{2}{c}{Width} \\
    \cmidrule(lr){2-3}\cmidrule(lr){4-5}\cmidrule(lr){6-7}
    $n$ & FM & NF & FM & NF & FM & NF \\
    \midrule
    $16$ & $0.100 \pm 0.001$ & $0.090 \pm 0.003$ & $0.035 \pm 0.002$ & $0.036 \pm 0.002$ & $0.080 \pm 0.006$ & $0.074 \pm 0.004$ \\
    $32$ & $0.074 \pm 0.001$ & $0.063 \pm 0.000$ & $0.022 \pm 0.000$ & $0.023 \pm 0.002$ & $0.055 \pm 0.002$ & $0.046 \pm 0.001$ \\
    $64$ & $0.064 \pm 0.001$ & $0.053 \pm 0.000$ & $0.016 \pm 0.000$ & $0.017 \pm 0.001$ & $0.053 \pm 0.008$ & $0.045 \pm 0.005$ \\
    $128$ & $0.059 \pm 0.002$ & $0.049 \pm 0.000$ & $0.013 \pm 0.000$ & $0.014 \pm 0.001$ & $0.059 \pm 0.010$ & $0.053 \pm 0.007$ \\
    $256$ & $0.059 \pm 0.003$ & $0.048 \pm 0.001$ & $0.012 \pm 0.000$ & $0.013 \pm 0.001$ & $0.066 \pm 0.011$ & $0.061 \pm 0.007$ \\
    $1024$ & $0.058 \pm 0.002$ & $0.047 \pm 0.002$ & $0.012 \pm 0.001$ & $0.014 \pm 0.002$ & $0.071 \pm 0.011$ & $0.069 \pm 0.008$ \\
    \bottomrule
  \end{tabular}
\end{table}

\subsection{Hyperparameters}

The embedder is the multi-stage set encoder of Algorithm~\ref{alg:multistage} with an
ISAB at every stage and weighted-mean pooling; the pooled output of the last stage is
the embedding $\embedding$. Both generator heads are conditioned on $\embedding$ by
concatenation: every MLP inside a head receives $\embedding$ appended to its input.
Table~\ref{tab:hparams-distributions} lists the hyperparameters of both heads,
together with the evaluation protocol.

\paragraph{Flow-matching head.} An MLP velocity field $v(x, t; \embedding)$ that
reads $x$ concatenated with a sinusoidal embedding of $t$. It is trained with the
conditional flow-matching loss on the linear path
$x_t = (1 - (1 - \sigma_{\min})\, t)\, z + t\, x$ from Gaussian noise $z$ to a sample
$x$, regressing the target velocity $x - (1 - \sigma_{\min})\, z$. Sampling integrates
the learned ODE from noise to data with the midpoint rule, and the density follows
from the continuous change-of-variables formula, accumulating the divergence of $v$
along the reverse path.

\paragraph{Normalizing-flow head.} A stack of masked coupling layers on a standard
normal base, each of which updates one coordinate given the other and then the
reverse, using the non-linear squared (NLSq) transform of \citet{ziegler2019latent},
$f(x) = a + bx + c / (1 + (dx + e)^2)$, whose five parameters are predicted by a
conditioned MLP. It is trained by maximum likelihood, and both sampling and density
evaluation take a single pass through the stack.

\paragraph{Evaluation.} Each held-out donut is embedded from $n$ of its samples. The
NLL is scored on fresh samples of the same donut, and a set of generated samples is
compared with the true donut by the sliced Wasserstein distance and by the recovery
error of its center, radius, and width, which are estimated from the generated
samples by the method of moments. The KDE and parametric fit baselines are fit to the same
$n$ samples.

\begin{table}[t]
  \centering
  \footnotesize
  \caption{Distribution-embedding architectures, hyperparameters, and evaluation protocol on the donut family.}
  \label{tab:hparams-distributions}
  \begin{tabular}{lrr}
    \toprule
    Hyperparameter & \multicolumn{2}{c}{Value} \\
    \midrule
    \multicolumn{3}{l}{\itshape Encoder (multi-stage ISAB)} \\
    Embedding size & \multicolumn{2}{c}{$16$} \\
    Stages & \multicolumn{2}{c}{$3$} \\
    Attention heads & \multicolumn{2}{c}{$4$} \\
    Inducing points & \multicolumn{2}{c}{$16$} \\
    Feed-forward hidden size & \multicolumn{2}{c}{$128$} \\
    \midrule
    \multicolumn{3}{l}{\itshape Generator head} \\
    Type & FM & NF \\
    Velocity MLP hidden size & $128$ & -- \\
    Velocity MLP layers & $4$ & -- \\
    Time-embedding size & $16$ & -- \\
    ODE steps (midpoint) & $100$ & -- \\
    $\sigma_{\min}$ & $10^{-4}$ & -- \\
    Coupling layers & -- & $8$ \\
    Parameter MLP hidden size & -- & $32$ \\
    Parameter MLP layers & -- & $3$ \\
    \midrule
    \multicolumn{3}{l}{\itshape Training (Adam)} \\
    Learning rate & \multicolumn{2}{c}{$0.001$} \\
    Warmup steps & \multicolumn{2}{c}{$0$} \\
    Batch size & \multicolumn{2}{c}{$64$} \\
    Samples per distribution & \multicolumn{2}{c}{$128$} \\
    Iterations & \multicolumn{2}{c}{$100{,}000$} \\
    \midrule
    \multicolumn{3}{l}{\itshape Evaluation} \\
    Held-out distributions & \multicolumn{2}{c}{$1{,}024$} \\
    NLL samples per distribution & \multicolumn{2}{c}{$512$} \\
    Generated samples per distribution & \multicolumn{2}{c}{$1{,}024$} \\
    SWD projections & \multicolumn{2}{c}{$1{,}000$} \\
    \bottomrule
  \end{tabular}
\end{table}

\section{Function Embeddings}
\label{app:function_embeddings}

\subsection{Families}

\paragraph{Polynomials.} Univariate polynomials
$f(x) = \sum_{k=0}^{d} c_k x^k$ of degree $d$ up to three, with coefficients drawn
uniformly and evaluated on a bounded interval. See Table~\ref{tab:hparams-function-families} for concrete ranges. Unlike the oscillatory ODE solutions
below, this family spans a wide range of output scales, from near-zero constants to
cubics reaching into the hundreds. We train the embedding over degrees $0$--$3$; the
differential operator is then trained over degrees $1$--$3$ only, since the derivative
of a degree-$0$ polynomial is identically zero and undefined under a relative-$L^2$
loss.

\paragraph{Damped harmonic oscillator.} A chain of $M$ unit masses on
a line, each tied to its rest position by an on-site spring of frequency $\omega$ and
to its neighbors by coupling springs, under uniform velocity damping $\gamma$. For a
single mass this reduces to $\ddot{x} + 2\gamma\dot{x} + \omega^2 x = 0$, whose
closed-form solution covers the under-, over-, and critically-damped regimes.
Stacking the $M$ positions and velocities into a state $z$ gives a linear system
$\dot{z} = A z$ with the closed-form solution $z(t) = \exp(A t)\, z_0$, so each
trajectory is a $2M$-dimensional state evolving in time. Varying
$M \in \{1, 2, 4, 8, 16\}$ sweeps the state dimension from $2$ to $32$, the axis of
the dimension-scaling result in Table~\ref{tab:fn_scaling}.

\paragraph{Lotka--Volterra.} The predator--prey system $\dot{x} = \alpha x -
\beta x y$, $\dot{y} = \delta x y - \gamma y$, with a two-dimensional state
$(x, y)$ tracking the prey and predator populations. It has no elementary
closed-form solution, so each trajectory is integrated numerically with a
Dormand--Prince solver implemented in \texttt{diffrax} \citep{kidger2021diffrax}. A function is
specified by its initial state $(x_0, y_0)$ and the four interaction rates
$(\alpha, \beta, \gamma, \delta)$, and the resulting trajectories are periodic
orbits of varying amplitude.

\begin{table}[t]
  \centering
  \footnotesize
  \caption{Function families: state dimension, and domain and parameter ranges.}
  \label{tab:hparams-function-families}
  \begin{tabular}{lrp{2.6in}}
    \toprule
    Family & State dim. & Domain and parameter ranges \\
    \midrule
    Polynomials & $1$ & $x \in [-4, 4]$, $c_k \in [-5, 5]$, $d \in [0, 3]$ \\
    Damped oscillator & $2M$ & $t \in [0, 10]$, $x_0, v_0 \in [-2, 2]$, $\gamma \in [0.3, 2.5]$, $\omega~\in~[0.5, 3]$, $k \in [0.5, 1.5]$ \\
    Lotka--Volterra & $2$ & $t \in [0, 10]$, $x_0, y_0 \in [0.5, 2]$, $\alpha, \beta, \gamma, \delta \in [0.8, 1.5]$ \\
    \bottomrule
  \end{tabular}
\end{table}

\subsection{Architectures \& Hyperparameters}

The embedder is the multi-stage set encoder of
Appendix~\ref{app:multistage_embedder} (Algorithm~\ref{alg:multistage}), whose
stages differ only in how each sample is combined with the running set-level
summary $c$. We use two such stage blocks, both built on the induced set attention
block (ISAB) of the Set Transformer \citep{lee2019st}. The two stage-block variants,
along with the decoder, operator, and training recipe, share a single set of
hyperparameters, listed in Table~\ref{tab:hparams-functions}.

\paragraph{Concatenation embedder.} Each stage concatenates the per-sample input
$x$, output $f(x)$, and (from the second stage on) the pooled summary $c$ into a
single vector, and passes the resulting set through an ISAB. This is the direct
analogue of the encoder used for distributions, with $c$ appended to every sample.

\paragraph{Dot-product embedder.} Each stage instead projects $x$, $f(x)$, and $c$
through separate MLPs and multiplies the three projections elementwise before
passing the product to the ISAB. The elementwise product lets the summary modulate
the per-sample features multiplicatively rather than enter as an extra input
channel, and we found it to yield a stronger encoder; it is the default in all
reported results. The two variants are compared in
Tables~\ref{tab:fn-embedding-families-st} and~\ref{tab:fn-embedding-families-stc}.

\paragraph{Decoder.} The decoder reconstructs $f$ at a query point $x$ by
single-head cross-attention. An MLP maps $x$ to a query vector, while two linear
projections expand the embedding $\embedding$ into a small fixed set of synthetic
key and value vectors; the query attends over the keys and returns the resulting
convex combination of values as $\hat{f}(x)$. Treating the embedding as a short
sequence of learned slots outperforms a plain MLP that concatenates $x$ and
$\embedding$ directly.

\paragraph{Operator networks.} Both operators are plain feedforward MLPs acting
directly in vector spaces. The differential operator maps a function's embedding to
the embedding of its derivative and is trained as a second stage on the frozen
embedding; the ODE solution operator maps a raw vector of initial conditions and
parameters to the embedding of the corresponding solution. In both cases the
decoder above expands the predicted embedding back into a function at arbitrary
query points.

\begin{table}[t]
  \centering
  \footnotesize
  \caption{Function-embedding architecture and training hyperparameters, shared by every family and both encoders.}
  \label{tab:hparams-functions}
  \begin{tabular}{lr}
    \toprule
    Hyperparameter & Value \\
    \midrule
    \multicolumn{2}{l}{\itshape Encoder (multi-stage ISAB)} \\
    Embedding size & $64$ \\
    Stages & $2$ \\
    Attention heads & $4$ \\
    Inducing points & $16$ \\
    Feed-forward / projection hidden size & $64$ \\
    Projection MLP layers & $2$ \\
    Activation & GELU \\
    \midrule
    \multicolumn{2}{l}{\itshape Decoder (cross-attention)} \\
    Query MLP hidden size & $128$ \\
    Query MLP layers & $2$ \\
    Query/key size & $48$ \\
    Key/value slots & $24$ \\
    Activation & GELU \\
    \midrule
    \multicolumn{2}{l}{\itshape Operator (MLP)} \\
    Hidden size & $256$ \\
    Hidden layers & $4$ \\
    Activation & GELU \\
    \midrule
    \multicolumn{2}{l}{\itshape Training (Adam)} \\
    Learning rate & $0.003$ \\
    Schedule & cosine decay \\
    Weight decay & $0$ \\
    Gradient clip & $1$ \\
    Loss & relative $L^2$ \\
    Batch size & $64$ \\
    Samples per function & $256$ \\
    Embedding iterations & $30{,}000$ \\
    Operator iterations & $30{,}000$ \\
    \midrule
    \multicolumn{2}{l}{\itshape Evaluation} \\
    Held-out functions & $1{,}024$ \\
    Samples per function & $256$ \\
    \bottomrule
  \end{tabular}
\end{table}

\subsection{Additional Results}

\begin{figure}[t]
    \centering
    \includegraphics[width=\linewidth]{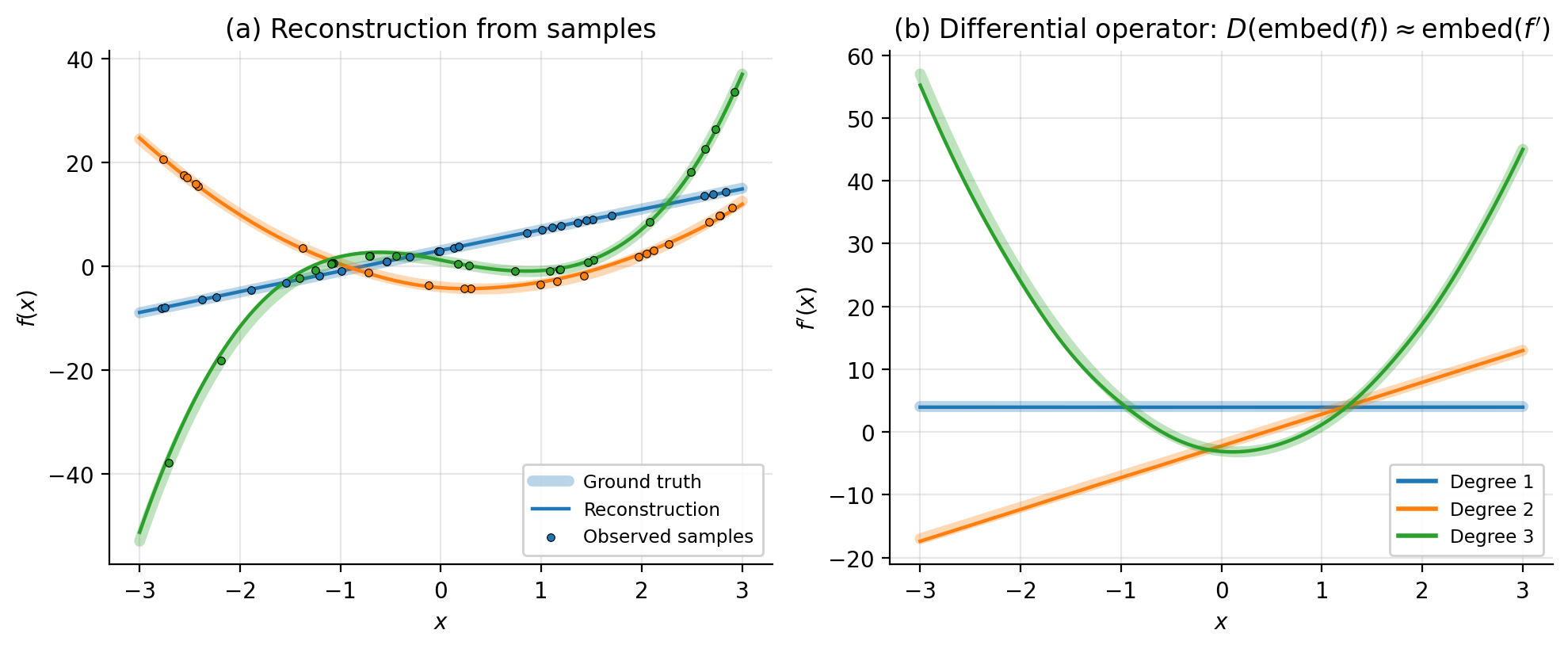}
    \caption{%
        \textbf{Function embeddings on the polynomial family.} Three polynomials of
        degree one, two, and three (color-coded). \textbf{(a)} Reconstruction from
        observed samples: the embedder maps the samples (dots) to an embedding from
        which the decoder reconstructs $f$ (solid line, ground truth shaded).
        \textbf{(b)} The differential operator maps each function's embedding to that
        of its derivative, $D(\mathrm{embed}(f)) \approx \mathrm{embed}(f')$, which the
        decoder then expands into $f'$ (solid line, ground truth shaded). A single
        embedding thus supports both reconstructing a function and differentiating it.%
    }
    \label{fig:fn-polynomials}
\end{figure}

Figure~\ref{fig:fn-polynomials} is the polynomial counterpart of the
Lotka--Volterra figure in the main text, for three polynomials of degree one
through three: the decoder reconstructs each $f$ from a small sample set (left),
while the differential operator maps its embedding to that of its derivative, which
the decoder in turn expands into $f'$ (right). Tables~\ref{tab:fn-embedding-families-st}
and~\ref{tab:fn-embedding-families-stc} give the full per-family reconstruction
error, broken out by polynomial degree and reported as RMSE, relative $L^2$, and
relative max-abs error, for the dot-product and concatenation encoders
respectively. Error grows with polynomial degree, as higher-degree polynomials span
a much wider range of output scales, and aggregated over each family the
dot-product encoder attains the lower error, consistent with our use of it as the
default in the main text.

\begin{table}
  \centering
  \small
  \caption{Embedding reconstruction error (inferred from a sample set) by family, Set Transformer encoder, concatenation fusion, mean $\pm$ SEM over 3 seeds. Rel Max is the relative max-abs error.}
  \label{tab:fn-embedding-families-stc}
  \begin{tabular}{@{}l ccc@{}}
    \toprule
    Family & RMSE & Rel $L^2$ & Rel Max \\
    \midrule
    Polynomials & $0.8596 \pm 0.0165$ & $0.0162 \pm 0.0007$ & $0.0184 \pm 0.0008$ \\
    \quad degree 0 & $0.0041 \pm 0.0002$ & $0.0010 \pm 0.0000$ & $0.0017 \pm 0.0001$ \\
    \quad degree 1 & $0.0660 \pm 0.0014$ & $0.0081 \pm 0.0004$ & $0.0100 \pm 0.0004$ \\
    \quad degree 2 & $0.4033 \pm 0.0246$ & $0.0182 \pm 0.0013$ & $0.0202 \pm 0.0014$ \\
    \quad degree 3 & $2.9649 \pm 0.0461$ & $0.0376 \pm 0.0011$ & $0.0415 \pm 0.0015$ \\
    Damped oscillator & $0.0114 \pm 0.0010$ & $0.0178 \pm 0.0007$ & $0.0191 \pm 0.0006$ \\
    Lotka--Volterra & $0.0087 \pm 0.0006$ & $0.0059 \pm 0.0003$ & $0.0092 \pm 0.0004$ \\
    \bottomrule
  \end{tabular}
\end{table}

\begin{table}
  \centering
  \small
  \caption{Embedding reconstruction error (inferred from a sample set) by family, Set Transformer encoder, dot-product fusion, mean $\pm$ SEM over 3 seeds. Rel Max is the relative max-abs error.}
  \label{tab:fn-embedding-families-st}
  \begin{tabular}{@{}l ccc@{}}
    \toprule
    Family & RMSE & Rel $L^2$ & Rel Max \\
    \midrule
    Polynomials & $0.5224 \pm 0.0682$ & $0.0143 \pm 0.0011$ & $0.0164 \pm 0.0017$ \\
    \quad degree 0 & $0.0055 \pm 0.0006$ & $0.0016 \pm 0.0001$ & $0.0030 \pm 0.0001$ \\
    \quad degree 1 & $0.0740 \pm 0.0048$ & $0.0103 \pm 0.0004$ & $0.0135 \pm 0.0011$ \\
    \quad degree 2 & $0.3786 \pm 0.0375$ & $0.0201 \pm 0.0019$ & $0.0221 \pm 0.0027$ \\
    \quad degree 3 & $1.6315 \pm 0.2362$ & $0.0251 \pm 0.0024$ & $0.0270 \pm 0.0032$ \\
    Damped oscillator & $0.0108 \pm 0.0018$ & $0.0173 \pm 0.0029$ & $0.0216 \pm 0.0034$ \\
    Lotka--Volterra & $0.0068 \pm 0.0003$ & $0.0047 \pm 0.0001$ & $0.0078 \pm 0.0000$ \\
    \bottomrule
  \end{tabular}
\end{table}

\section{Graph Embeddings}
\label{app:graph_embeddings}

\subsection{Families}

\paragraph{Random trees.} Uniformly random labeled trees on $n$ vertices, drawn
through Pr\"ufer sequences: a uniformly random length-$(n-2)$ string over the $n$
vertex labels decodes to a distinct labeled tree, so every labeled tree is equally
likely. The graphs are unweighted, and we evaluate them at $n \in \{10, 20\}$.

\paragraph{Scale-free networks.} Barab\'asi--Albert graphs \citep{albert2002statistical} grown by preferential
attachment: vertices are added one at a time, and each new vertex attaches to
$m = 2$ earlier vertices chosen with probability proportional to their current
degree. This yields a heavy-tailed degree distribution with a few high-degree hubs,
which are the vertices reconstruction struggles with most. The graphs are
unweighted, again at $n \in \{10, 20\}$.

\paragraph{Gridworlds.} The only weighted family. Each graph starts from a fixed
$4 \times 4$ grid ($n = 16$) and removes up to three of its edges
uniformly at random, subject to the remaining graph staying connected. Every
surviving edge carries a positive weight drawn uniformly, and edges are sampled with
frequency proportional to their weight, so the family tests reconstruction of a
weighted adjacency rather than a binary one.

\subsection{Architectures \& Hyperparameters}

\label{app:graph_architectures}

The encoder is the multi-stage set encoder of Appendix~\ref{app:multistage_embedder},
built on ISABs and applied to the sampled edges, each given as the pair of one-hot
labels of its endpoints. Table~\ref{tab:hparams-graphs} lists the hyperparameters;
across families only the number of representation dimensions and the number of edges
sampled per graph change. Each sampled edge set is split in half between producing
the embedding and scoring the decoder, and every configuration is trained with three
seeds.

\paragraph{Decoder.} The decoder is the ALLO objective of Section~\ref{subsec:allo}
with both of its trainable parts conditioned on the embedding. An MLP maps a vertex's
one-hot label, concatenated with $\embedding$, to its $d$-dimensional representation,
and a second MLP predicts each dual variable $\beta_{jk}$ from $\embedding$ and the
index pair $(j, k)$. The diagonal duals provide the eigenvalue estimates; the
representation used downstream drops the constant eigenvector and rescales every
remaining coordinate by the inverse square root of its eigenvalue. The augmented
Lagrangian carries the penalty coefficient $b$ and an $L_2$ regularizer on the dual
variables, and its saddle point is optimized with optimistic Adam.

\paragraph{Reconstruction heads.} The medians network is an MLP that maps a vertex's
one-hot label, concatenated with $\embedding$, through a softplus to $m_u$, the
predicted median distance from vertex $u$ to its neighbors. It is trained with the
mean absolute error discussed in Section~\ref{sec:graph_embeddings} and predicts an edge between
$u$ and $v$ whenever $d(u, v) \leq m_u + m_v$. The degree oracle instead connects
each vertex to its $\deg(u)$ nearest vertices, thresholding halfway between the
$\deg(u)$-th and the next nearest distance, with the threshold of a pair taken as the
mean of its endpoints' thresholds. For the weighted gridworld family, the transition
network is an MLP that scores every ordered vertex pair from its representation
distance and $\embedding$; a global softmax over all pairs, excluding self-loops,
turns the scores into a normalized symmetric adjacency, trained by the negative
log-likelihood of the held-out sampled edges. Since edges are sampled with frequency
proportional to their weight, this recovers the true adjacency up to its overall
scale, which frequency sampling cannot identify.

\paragraph{Evaluation.} Each held-out graph is embedded from its sampled edges, and
the geometry is scored by ranking every vertex pair by representation distance: we
report the ROC-AUC and the average precision (AP) of the resulting edge detector.
Reconstruction is scored by the precision, recall, and F1 of the predicted adjacency
against the true one, with counts pooled over all held-out graphs, for both the degree
oracle and the medians network. On gridworld, the transition head's adjacency is
compared with the true normalized adjacency by Spearman rank correlation, over the
true edges and over all vertex pairs (rescaled by the ceiling that the tied non-edges
impose), and by the mean absolute error of the off-diagonal weights.

\begin{table}[t]
  \centering
  \footnotesize
  \caption{Graph-embedding architectures and hyperparameters per family.}
  \label{tab:hparams-graphs}
  \begin{tabular}{lrrrrr}
    \toprule
    Hyperparameter & Trees $10$ & Trees $20$ & SF $10$ & SF $20$ & Gridworld \\
    \midrule
    \multicolumn{6}{l}{\itshape Encoder (multi-stage ISAB)} \\
    Embedding size & \multicolumn{5}{c}{$128$} \\
    Stages & \multicolumn{5}{c}{$2$} \\
    Attention heads & \multicolumn{5}{c}{$4$} \\
    Inducing points & \multicolumn{5}{c}{$32$} \\
    Feed-forward hidden size & \multicolumn{5}{c}{$64$} \\
    \midrule
    \multicolumn{6}{l}{\itshape Decoder} \\
    Representation dims $d$ & $8$ & $16$ & $8$ & $16$ & $12$ \\
    Eigenfunction MLP hidden size & \multicolumn{5}{c}{$1{,}024$} \\
    Eigenfunction MLP layers & \multicolumn{5}{c}{$3$} \\
    Dual-variable MLP hidden size & \multicolumn{5}{c}{$32$} \\
    Dual-variable MLP layers & \multicolumn{5}{c}{$2$} \\
    Penalty coefficient $b$ & \multicolumn{5}{c}{$2$} \\
    Dual regularization & \multicolumn{5}{c}{$0.003$} \\
    \midrule
    \multicolumn{6}{l}{\itshape Reconstruction head (MLP)} \\
    Hidden size & \multicolumn{5}{c}{$64$} \\
    Hidden layers & \multicolumn{5}{c}{$3$} \\
    \midrule
    \multicolumn{6}{l}{\itshape Embedding training (optimistic Adam)} \\
    Learning rate & \multicolumn{5}{c}{$0.001$} \\
    Schedule & \multicolumn{5}{c}{cosine decay} \\
    Gradient clip & \multicolumn{5}{c}{$1$} \\
    Batch size & \multicolumn{5}{c}{$64$} \\
    Samples per graph & $128$ & $256$ & $128$ & $256$ & $256$ \\
    Iterations & \multicolumn{5}{c}{$100{,}000$} \\
    \midrule
    \multicolumn{6}{l}{\itshape Head training (Adam)} \\
    Learning rate & \multicolumn{5}{c}{$0.001$} \\
    Schedule & \multicolumn{5}{c}{constant} \\
    Batch size & \multicolumn{5}{c}{$64$} \\
    Samples per graph & $128$ & $256$ & $128$ & $256$ & $256$ \\
    Iterations & \multicolumn{5}{c}{$50{,}000$} \\
    \midrule
    \multicolumn{6}{l}{\itshape Evaluation} \\
    Held-out graphs & \multicolumn{5}{c}{$1{,}024$} \\
    Samples per graph & $128$ & $256$ & $128$ & $256$ & $256$ \\
    \bottomrule
  \end{tabular}
\end{table}

\subsection{Additional Results}

Table~\ref{tab:graph-appendix} extends Table~\ref{tab:graph_families} with the average
precision of the geometry, the full precision and recall of both reconstruction heads,
and the weighted gridworld family. The medians network reconstructs with a recall near
one but a low precision on the $n = 20$ families, the signature of the over-prediction
discussed in the main text, whereas the degree oracle keeps the two balanced. On
gridworld the geometry separates edges from non-edges nearly as well as on the
$n = 10$ families, and the transition head ranks the true edge weights with a
Spearman correlation of about $0.76$ over the edges and $0.93$ over all vertex
pairs, after rescaling for the ceiling that the tied non-edges impose.

\begin{table}[t]
  \centering
  \footnotesize
  \setlength{\tabcolsep}{4pt}
  \caption{Graph-embedding quality per family, all metrics (mean $\pm$ SEM over 3 seeds).}
  \label{tab:graph-appendix}
  \begin{tabular}{lrrrrr}
    \toprule
    Metric & Trees $10$ & Trees $20$ & SF $10$ & SF $20$ & Gridworld \\
    \midrule
    \multicolumn{6}{l}{\itshape Geometry} \\
    ROC-AUC & $0.993 \pm 0.001$ & $0.953 \pm 0.006$ & $0.980 \pm 0.001$ & $0.847 \pm 0.010$ & $0.958 \pm 0.002$ \\
    AP & $0.978 \pm 0.002$ & $0.793 \pm 0.016$ & $0.956 \pm 0.001$ & $0.625 \pm 0.018$ & $0.907 \pm 0.001$ \\
    \midrule
    \multicolumn{6}{l}{\itshape Reconstruction (degree oracle)} \\
    Precision & $0.984 \pm 0.001$ & $0.806 \pm 0.023$ & $0.949 \pm 0.002$ & $0.649 \pm 0.021$ & \textemdash \\
    Recall & $0.982 \pm 0.001$ & $0.758 \pm 0.032$ & $0.967 \pm 0.002$ & $0.580 \pm 0.023$ & \textemdash \\
    F1 & $0.983 \pm 0.001$ & $0.781 \pm 0.028$ & $0.958 \pm 0.002$ & $0.613 \pm 0.022$ & \textemdash \\
    \midrule
    \multicolumn{6}{l}{\itshape Reconstruction (medians)} \\
    Precision & $0.879 \pm 0.009$ & $0.299 \pm 0.059$ & $0.658 \pm 0.003$ & $0.199 \pm 0.006$ & \textemdash \\
    Recall & $0.970 \pm 0.001$ & $0.957 \pm 0.011$ & $0.997 \pm 0.000$ & $0.994 \pm 0.001$ & \textemdash \\
    F1 & $0.922 \pm 0.005$ & $0.449 \pm 0.064$ & $0.793 \pm 0.002$ & $0.331 \pm 0.008$ & \textemdash \\
    \midrule
    \multicolumn{6}{l}{\itshape Weight reconstruction} \\
    Spearman (edges) & \textemdash & \textemdash & \textemdash & \textemdash & $0.756 \pm 0.003$ \\
    Spearman (all pairs) & \textemdash & \textemdash & \textemdash & \textemdash & $0.926 \pm 0.003$ \\
    MAE ($\times 10^{-3}$) & \textemdash & \textemdash & \textemdash & \textemdash & $1.700$ \\
    \bottomrule
  \end{tabular}
\end{table}

\section{Source Code}
\label{app:source_code}

The entire framework is implemented in JAX \citep{jax2018github}. The source code will be publicly available on GitHub upon publication.

\end{document}